\documentclass[a4paper,fleqn]{cas-dc}

\usepackage[authoryear,longnamesfirst]{natbib}
\usepackage{graphicx}%
\usepackage{multirow}%
\usepackage{amsmath,amssymb,amsfonts}%
\usepackage{amsthm}%
\usepackage{mathrsfs}%
\usepackage[title]{appendix}%
\usepackage{xcolor}%
\usepackage{textcomp}%
\usepackage{manyfoot}%
\usepackage{booktabs}%
\usepackage{algorithm}%
\usepackage{algorithmicx}%
\usepackage{algpseudocode}%
\usepackage{listings}%
\usepackage{booktabs}
\usepackage{longtable}
\usepackage{placeins}

\def\tsc#1{\csdef{#1}{\textsc{\lowercase{#1}}\xspace}}
\tsc{WGM}
\tsc{QE}

\begin{document}
\let\WriteBookmarks\relax
\def\floatpagepagefraction{1}
\def\textpagefraction{.001}

\shorttitle{}    

\shortauthors{}  

\title [mode = title]{Pre-Deployment Robustness Stress Testing for CT Segmentation Systems Using Clinically Motivated Multi-Corruption Augmentation}  

\author[1]{CholMin Kang}
\cormark[1]
\ead{kcholmin@gmail.com}

\author[2]{Jonghyun Chung}
\ead{johnchung@google.com}

\author[2]{Amanpreet Kaur}
\ead{amanpk@google.com}

\author[2]{Nagesh Gulkotwar}
\ead{ngulkotwar@google.com}

\author[2]{Aarthi Sivasankaran}
\ead{asivasankaran@google.com}

\affiliation[1]{organization={Seoul National University},
            addressline={1 Gwanak-ro, Gwanak-gu}, 
            city={Seoul},
            country={Korea}}
\affiliation[2]{organization={Google Inc.},
            addressline={1600 Amphitheatre Parkway}, 
            city={Mountain View},
            postcode={94043}, 
            state={California},
            country={United States}}

\cortext[1]{Corresponding author}


\begin{abstract}
Deep learning-based CT segmentation systems often achieve high accuracy on clean benchmark images, but their performance may degrade under heterogeneous clinical imaging conditions such as noise, resolution loss, contrast variation, intensity shift, and artifacts. This instability can limit reliable deployment in real-world medical imaging workflows. We propose Robustness via Augmented Multi-corruption Pipeline (RAMP), a robustness-oriented augmentation framework for CT segmentation. RAMP combines anatomically constrained spatial perturbations, CT intensity transformations, and stochastic multi-corruption composition to expose models to clinically plausible image degradation during training. Across two CT segmentation evaluation settings, RAMP achieved the strongest corrupted-image performance and the smallest clean-to-corrupted robustness gap. In the five-organ noisy evaluation benchmark, RAMP improved mean corrupted Dice from 0.610 to 0.753 and reduced the robustness gap from 0.264 to 0.064 compared with the nnU-Net baseline. In Abdomen1K, RAMP improved mean corrupted Dice from 0.633 to 0.789 and reduced the robustness gap from 0.290 to 0.070. Although RAMP did not achieve the highest clean-image Dice, it substantially mitigated worst-case segmentation collapse under severe image degradation. These results suggest that multi-corruption augmentation can serve as a practical pre-deployment strategy for improving the reliability of CT segmentation systems in heterogeneous clinical environments.\nocite{*}
\end{abstract}




\maketitle

\section{Introduction}\label{sec1}
Deep learning-based medical image segmentation has become an important component of modern computer-assisted diagnosis, treatment planning, quantitative imaging, and automated clinical workflow systems. In CT imaging, segmentation models are used to delineate anatomical organs and structures for volumetric measurement, radiomics analysis, surgical planning, radiation therapy planning, and large-scale retrospective imaging studies. Convolutional neural networks have played a central role in this progress, beginning with encoder--decoder architectures such as U-Net and extending to volumetric models such as 3D U-Net and V-Net \cite{ronneberger2015unet,cicek20163dunet,milletari2016vnet}. In addition to convolutional architectures, recent studies have explored transformer-based and pretrained 3D representations for volumetric medical image segmentation, further broadening the design space of segmentation systems \cite{hatamizadeh2022unetr,zhou2021modelsgenesis}.
More recently, self-configuring frameworks such as nnU-Net have demonstrated strong and reproducible performance across diverse biomedical segmentation tasks \cite{isensee2021nnunet}. These advances have made deep learning segmentation systems increasingly practical for medical imaging workflows.

Despite this progress, high performance on clean benchmark datasets does not necessarily imply reliable performance in real-world clinical deployment. Deep learning models for medical imaging are often trained and evaluated on curated datasets that may not fully capture the heterogeneity of routine clinical data \cite{litjens2017survey}. CT images acquired across different institutions, scanners, acquisition protocols, reconstruction kernels, slice thicknesses, contrast phases, and dose settings may exhibit substantial variation in appearance. Such variation can lead to domain shift between the training distribution and the deployment environment. Domain shift is a well-recognized challenge in medical image analysis and can substantially reduce the reliability of machine learning models when applied outside their original development setting \cite{guan2022domainadaptation,zech2018variable}. This concern is consistent with broader discussions on clinical AI deployment, where model performance, reporting transparency, workflow integration, and prospective evaluation are considered essential for safe translation into practice \cite{topol2019highperformance,yu2018aihealthcare,nagendran2020artificial,mongan2020claim,vasey2022decideai}. For segmentation systems, this problem is particularly important because degraded segmentation masks can affect downstream quantitative measurements and clinical decision-support pipelines.

Large-scale segmentation datasets and benchmarks have improved the development and comparison of medical segmentation algorithms across anatomical regions, imaging modalities, and clinical tasks \cite{menze2015brats,antonelli2022msd,ma2022abdomenct1k,ji2022amos,wasserthal2023totalsegmentator}. The Medical Segmentation Decathlon provided a multi-task benchmark for evaluating generalizable biomedical segmentation methods \cite{antonelli2022msd}. AbdomenCT-1K further highlighted that abdominal organ segmentation remains challenging under multi-center, multi-phase, multi-vendor, and multi-disease conditions, even when state-of-the-art methods achieve strong performance on conventional test sets \cite{ma2022abdomenct1k}. TotalSegmentator demonstrated the feasibility of robust multi-structure segmentation across whole-body CT images, further emphasizing the clinical value of scalable segmentation systems \cite{wasserthal2023totalsegmentator}. However, most segmentation benchmarks still primarily report performance on clean or naturally acquired images. They do not always explicitly evaluate how models behave under controlled degradation such as noise, blur, resolution loss, intensity shift, contrast variation, and artifacts.

Robustness evaluation is especially important for medical AI systems because real-world imaging conditions are rarely ideal. In routine CT workflows, image quality can be affected by low-dose acquisition, patient motion, reconstruction artifacts, variable contrast enhancement, local intensity non-uniformity, and differences in scanner protocols. A model that performs well on clean validation images may fail when exposed to such perturbations. This phenomenon is consistent with broader observations in machine learning: models can exploit shortcuts that perform well under standard benchmark conditions but fail under distribution shift or more challenging testing conditions \cite{geirhos2020shortcut}. Related work has also shown that medical image analysis models may exhibit clinically meaningful failures on hidden subgroups or under small input perturbations, highlighting the need for evaluation protocols that go beyond average test-set performance \cite{oakdenrayner2020hidden,goodfellow2015explaining,finlayson2019adversarial}. In computer vision, common corruption benchmarks have been proposed to evaluate model robustness under realistic non-adversarial perturbations \cite{hendrycks2019benchmarking}. A similar perspective is needed for medical image segmentation, where robustness should be evaluated not only by average clean-image accuracy but also by stability under clinically plausible image degradation.

Data augmentation is one of the most widely used strategies for improving generalization in deep learning, including medical image segmentation \cite{shorten2019survey,perezgarcia2021torchio}. Standard augmentation pipelines often include spatial transformations, intensity perturbations, noise injection, and elastic deformation. Beyond hand-designed transformations, automated or robustness-oriented augmentation methods such as AutoAugment and AugMix have shown that composition and stochasticity in augmentation policies can improve model generalization and robustness to distribution shift in natural image settings \cite{cubuk2019autoaugment,hendrycks2020augmix}. These techniques can improve invariance to limited variations in anatomy and image appearance. However, many augmentation strategies are optimized primarily for clean validation performance rather than robustness under severe image degradation. Moreover, real clinical degradation is often compound rather than isolated. A CT scan may simultaneously exhibit increased noise, reduced resolution, altered contrast, and reconstruction artifacts. Training with only single or mild perturbations may therefore be insufficient to prevent model collapse under more challenging deployment conditions.

In this study, we frame CT segmentation robustness as a pre-deployment medical systems problem. Rather than focusing solely on maximizing clean-image Dice score, we aim to reduce the clean-to-corrupted performance gap and prevent severe segmentation failure under degraded imaging conditions. This framing is aligned with the broader need for reliable and responsibly deployed medical AI systems, where performance should be assessed under conditions that approximate real clinical use rather than only under ideal benchmark settings \cite{kelly2019key,wiens2019donoharm}. For segmentation-based medical systems, robustness is not merely a methodological property; it is a practical requirement for maintaining stable downstream measurements, workflow reliability, and clinical usability.

To address this problem, we propose Robustness via Augmented Multi-corruption Pipeline (RAMP), a stochastic multi-corruption augmentation framework for robust CT segmentation. RAMP integrates three components. First, anatomically constrained spatial perturbations are applied to improve geometric invariance while preserving label consistency. Second, CT intensity transformations simulate variations in windowing, local contrast, histogram shape, and sharpness. Third, a random multi-corruption sampler applies multiple degradation operators to each training image, including noise, blur, motion-related degradation, low-frequency intensity variation, global intensity shift, contrast reduction, resolution degradation, stripe or banding artifacts, and compound corruption. By exposing the model to a broad perturbation space during training, RAMP is designed to improve segmentation stability under heterogeneous imaging conditions.

The central hypothesis of this work is that stochastic multi-corruption training can reduce robustness gaps under degraded CT imaging conditions, even if it does not maximize clean-image performance. This hypothesis reflects an important clinical trade-off. For deployment-oriented medical AI systems, a slightly lower clean-image Dice score may be acceptable if the model substantially reduces worst-case failure and maintains non-trivial performance under severe degradation. Therefore, we evaluate segmentation models using not only clean Dice but also mean corrupted Dice, clean-to-corrupted robustness gap, and worst-case corrupted Dice. These metrics provide a more deployment-relevant assessment of segmentation reliability than clean-image performance alone.

We evaluate RAMP across two CT segmentation settings and compare it with conventional augmentation strategies, including baseline nnU-Net training, intensity-focused augmentation, geometric augmentation, and artifact-focused augmentation. Our results show that RAMP consistently provides the strongest corrupted-image performance and the smallest robustness gap. In the five-organ noisy evaluation benchmark, RAMP improves mean corrupted Dice from 0.610 to 0.753 compared with the nnU-Net baseline and reduces the robustness gap from 0.264 to 0.064. In AbdomenCT-1K, RAMP improves mean corrupted Dice from 0.633 to 0.789 and reduces the robustness gap from 0.290 to 0.070. Although several conventional augmentation strategies achieve higher clean-image Dice, they exhibit severe degradation under high-severity corruptions, with worst-case performance approaching zero. In contrast, RAMP preserves non-trivial segmentation performance under severe image degradation.

The contributions of this study are threefold. First, we formulate robustness evaluation for CT segmentation as a pre-deployment medical systems problem, emphasizing the need to assess model stability under heterogeneous imaging conditions. Second, we propose RAMP, a stochastic multi-corruption augmentation framework designed to improve robustness against noise, intensity variation, resolution degradation, artifacts, and compound perturbations. Third, we demonstrate that RAMP substantially reduces clean-to-corrupted performance gaps and mitigates worst-case segmentation collapse across two CT segmentation evaluation settings. These findings suggest that robustness-oriented augmentation and stress testing may improve the reliability of CT segmentation systems before deployment in real-world clinical workflows.
\section{Proposed Method}

\subsection{Overview}

\begin{figure*}[t]
    \centering
    \includegraphics[width=0.99\textwidth, keepaspectratio]{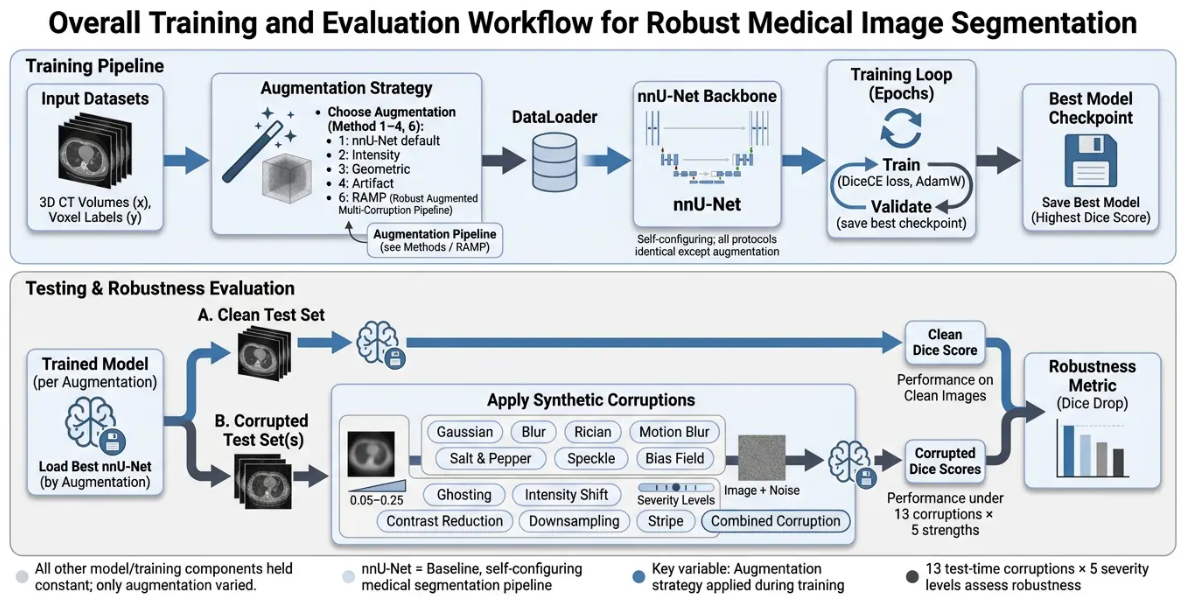}
    \caption{Overall training and evaluation workflow. Segmentation models were trained using different augmentation strategies and evaluated under both clean and corrupted CT imaging conditions. The proposed RAMP framework was designed to improve robustness under heterogeneous image degradation rather than optimize only for clean-image Dice performance.}
    \label{fig:overall_train_test}
\end{figure*}

\begin{figure*}[t]
    \centering
    \includegraphics[width=0.99\textwidth]{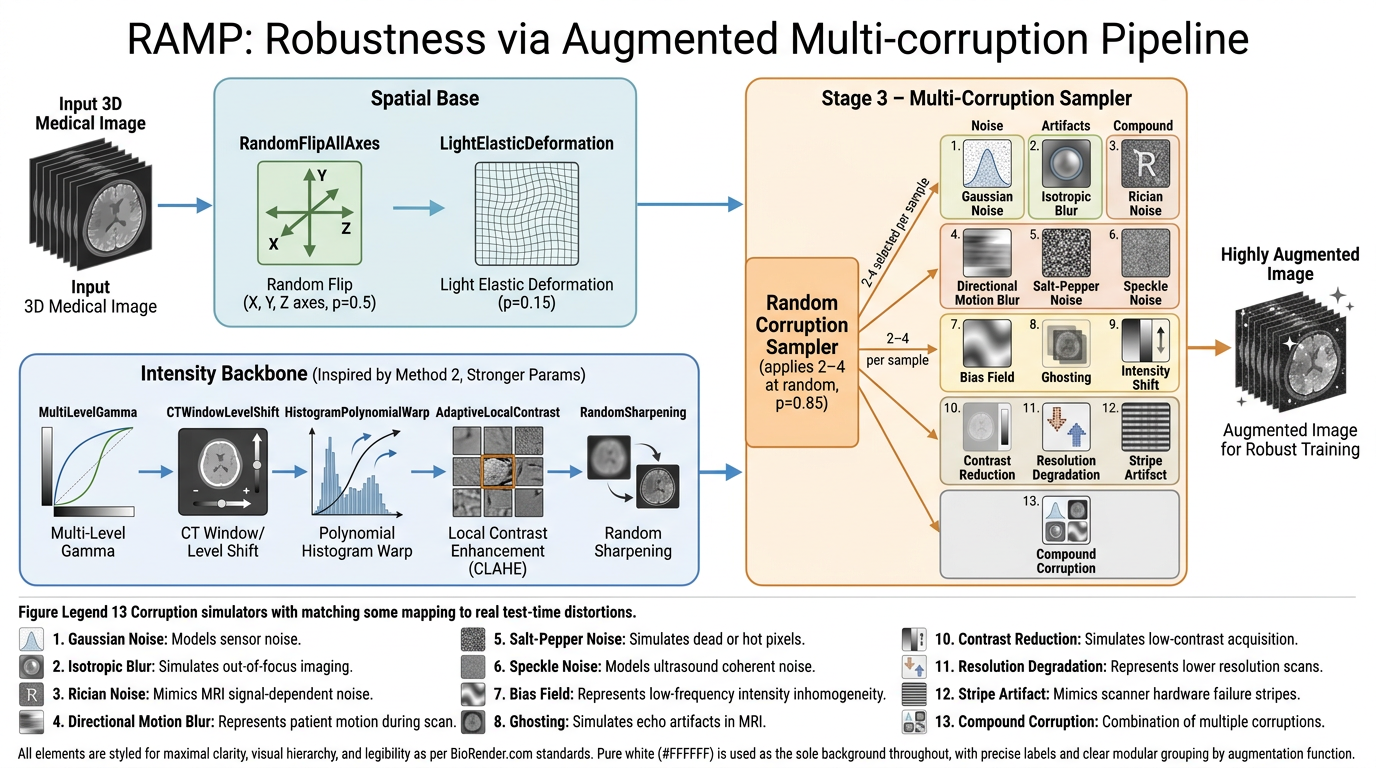}
    \caption{Overview of the proposed RAMP augmentation framework. RAMP consists of anatomically constrained spatial perturbation, CT intensity transformation, and stochastic multi-corruption composition. During training, multiple image degradation operators are randomly selected and applied to each image to simulate heterogeneous clinical imaging conditions.}
    \label{fig:main_augmentation}
\end{figure*}

We propose Robustness via Augmented Multi-corruption Pipeline (RAMP), a robustness-oriented augmentation framework for CT segmentation systems. The objective of RAMP is not to maximize clean-image segmentation performance alone, but to reduce segmentation instability under heterogeneous imaging conditions that may occur during clinical deployment. The overall training and evaluation workflow is illustrated in Fig.~\ref{fig:overall_train_test}, and the detailed augmentation pipeline is shown in Fig.~\ref{fig:main_augmentation}.

Deep learning-based segmentation models are commonly evaluated on clean held-out datasets, but such evaluation may not fully capture the distribution shifts encountered in routine clinical imaging. Previous work has shown that convolutional neural networks can be sensitive to common corruptions and non-adversarial perturbations \cite{hendrycks2019benchmarking}. In medical imaging, domain shift caused by differences in institutions, scanners, acquisition protocols, and patient populations is also a major challenge for reliable deployment \cite{zech2018variable,guan2022domainadaptation}. RAMP was therefore designed as a pre-deployment robustness strategy that exposes segmentation models to a broad family of clinically plausible perturbations during training.

\subsection{Baseline Segmentation Framework}

We used nnU-Net as the baseline segmentation framework because it provides a strong and reproducible self-configuring pipeline for biomedical image segmentation \cite{isensee2021nnunet}. The choice of a standardized segmentation framework is consistent with prior efforts to improve reproducibility and modularity in medical image deep learning pipelines \cite{gibson2018niftynet,perezgarcia2021torchio,cardoso2022monai}. nnU-Net automatically adapts key components of the segmentation pipeline, including preprocessing, network configuration, training schedule, and post-processing, based on dataset properties. This makes it a suitable reference framework for evaluating the effect of augmentation strategies without requiring manual architecture tuning.

Let $x \in \mathbb{R}^{1 \times D \times H \times W}$ denote a 3D CT volume and $y \in \{0,\ldots,C\}^{D \times H \times W}$ denote the corresponding voxel-level segmentation label map. A segmentation network $f_{\theta}$ is trained to predict $\hat{y}=f_{\theta}(x)$ by minimizing the segmentation loss between the predicted probability map and the ground-truth mask. In this study, the primary methodological variable was the augmentation strategy applied during training. All compared methods used the same segmentation backbone and training protocol, while differing only in the augmentation pipeline.

\subsection{RAMP Augmentation Pipeline}

RAMP consists of three sequential stages: anatomically constrained spatial perturbation, CT intensity transformation, and stochastic multi-corruption composition. The purpose of this pipeline is to expose the segmentation model to a broad range of clinically plausible image degradation patterns during training while preserving the anatomical consistency of the corresponding segmentation labels. Unlike a single-perturbation augmentation strategy, RAMP was designed to simulate compound image degradation, where multiple sources of variation may occur simultaneously in clinical CT imaging.

All CT images were first clipped to a task-appropriate Hounsfield unit range and normalized to the interval $[0,1]$. RAMP was applied after this normalization step, meaning that all intensity and corruption operations were performed in normalized intensity space rather than directly in raw Hounsfield unit space. After each intensity or corruption operation, image values were clipped to $[0,1]$ to prevent out-of-range values. Segmentation labels were transformed only by spatial operations and were not modified by intensity or corruption operations.

The first stage applies light spatial augmentation to improve geometric invariance while avoiding unrealistic anatomical distortion. Random flipping was applied independently along the depth, height, and width axes with a probability of 0.50 per axis. Mild elastic deformation was applied with a probability of 0.15 using Gaussian-smoothed random displacement fields. Linear interpolation was used for CT images, whereas nearest-neighbor interpolation was used for segmentation masks to preserve discrete label values. In the present experiments, the segmentation task used organ-level labels. If laterality-specific labels are retained in future experiments, such as separate left and right kidney labels, left-right flipping should either be disabled or accompanied by appropriate label swapping.

The second stage applies CT intensity transformations to simulate variation in image appearance caused by differences in windowing, reconstruction, local contrast, and intensity distribution. This stage includes two-step gamma transformation, CT window-level shift, monotonic histogram polynomial warping, adaptive local contrast adjustment, and random sharpening. Each transformation is applied independently according to its own probability. These operations were selected to represent intensity-level heterogeneity commonly observed across CT acquisition and reconstruction protocols.

The third stage applies stochastic multi-corruption composition. With probability $p_{\mathrm{RAMP}}=0.85$, RAMP randomly samples between two and four corruption operators from the corruption pool and applies them sequentially to the training image. The corruption operators are sampled without replacement for each image. If stochastic multi-corruption composition is not applied, the image may still receive spatial and intensity augmentations from the previous stages. The corruption pool includes additive Gaussian noise, isotropic blur, Rician-like noise, directional motion blur, salt-and-pepper noise, speckle noise, low-frequency intensity non-uniformity, ghosting-like artifact, global intensity shift, contrast reduction, resolution degradation, stripe or banding artifact, and compound corruption.

Formally, let $x \in \mathbb{R}^{1 \times D \times H \times W}$ denote a normalized CT image and $y$ denote the corresponding segmentation label. Let $\mathcal{A}_{s}$ denote the spatial augmentation operator, $\mathcal{A}_{i}$ denote the sequence of independently sampled intensity transformations, and $\mathcal{C}=\{c_1,c_2,\ldots,c_M\}$ denote the corruption operator pool. For each training sample, RAMP first samples a Bernoulli variable $b \sim \mathrm{Bernoulli}(p_{\mathrm{RAMP}})$. If $b=1$, the number of corruption operators is sampled as $k \sim \mathcal{U}\{2,3,4\}$, and $k$ operators are sampled without replacement from $\mathcal{C}$. The augmented image is generated as:
\begin{equation}
    \tilde{x}
    =
    c_{j_k} \circ c_{j_{k-1}} \circ \cdots \circ c_{j_1}
    \circ \mathcal{A}_{i}
    \circ \mathcal{A}_{s}(x),
\end{equation}
where $\{c_{j_1}, \ldots, c_{j_k}\} \subset \mathcal{C}$ are the selected corruption operators. The corresponding label is transformed as:
\begin{equation}
    \tilde{y} = \mathcal{A}_{s}(y),
\end{equation}
because intensity and corruption operators affect only the image appearance and not the anatomical ground truth.

The training-time corruption severities were sampled continuously from the parameter ranges listed below. They were not tied to a single discrete test severity level. This design was intended to expose the model to a broad perturbation space rather than optimize it for a specific test corruption setting. During evaluation, corruption severities were applied in a controlled manner to assess model behavior under increasing degradation.

\begingroup
\scriptsize
\setlength{\tabcolsep}{2.5pt}
\renewcommand{\arraystretch}{1.15}

\begin{table*}[t]
\centering
\scriptsize
\caption{Default parameters of the RAMP spatial and intensity augmentation stages. All operations are applied in normalized $[0,1]$ intensity space unless otherwise specified.}
\label{tab:ramp_stage_params}
\resizebox{\textwidth}{!}{
\begin{tabular}{llp{4.2cm}p{7.2cm}}
\hline
Stage & Operation & Default parameters & Implementation details \\
\hline
Spatial & Random flip & $p=0.50$ per axis & Applied independently along depth, height, and width axes. The same flip is applied to the image and segmentation mask. \\

Spatial & Elastic deformation & $p=0.15$, $\alpha=80.0$, $\sigma=10.0$ & Gaussian-smoothed random displacement fields. Image interpolation uses linear interpolation; mask interpolation uses nearest-neighbor interpolation. \\

Intensity & Multi-level gamma & $p=0.45$, $\gamma_1 \in [0.60,1.60]$, $\gamma_2 \in [0.70,1.30]$ & Two sequential gamma transformations are applied to the normalized image. \\

Intensity & CT window-level shift & $p=0.45$, center shift $\in [-0.15,0.15]$, width scale $\in [0.70,1.30]$ & Simulates variation in CT windowing and intensity scaling. \\

Intensity & Histogram polynomial warp & $p=0.40$, degree $=3$, coefficient std $=0.20$ & Uses positive normalized polynomial coefficients to generate a monotonic intensity remapping. \\

Intensity & Adaptive local contrast & $p=0.35$, grid $=4 \times 4 \times 4$, max blend $=0.35$ & Applies patch-wise local normalization and blends it with the original image using a random blend factor. \\

Intensity & Random sharpening & $p=0.30$, Gaussian $\sigma \in [0.5,2.5]$, strength $\alpha \in [0.2,1.5]$ & Uses unsharp masking to simulate variation in edge sharpness and reconstruction appearance. \\

Composition & Multi-corruption sampler & $p_{\mathrm{RAMP}}=0.85$, $k \in \{2,3,4\}$ & Samples $k$ corruption operators without replacement and applies them sequentially to the image. \\
\hline
\end{tabular}
}
\end{table*}
\endgroup

\begingroup
\begin{table*}[t]
\centering
\scriptsize
\caption{Default parameters of the RAMP corruption operator pool. Corruptions are applied only to normalized CT images and not to segmentation labels.}
\label{tab:ramp_corruption_params}
\resizebox{\textwidth}{!}{
\begin{tabular}{lp{4.2cm}p{8.5cm}}
\hline
Corruption operator & Default parameters & Implementation details \\
\hline
Gaussian noise & $\sigma \in [0.02,0.20]$ & Adds independent Gaussian noise to each voxel and clips the result to $[0,1]$. \\

Isotropic blur & Gaussian $\sigma \in [0.5,2.5]$ & Applies Gaussian smoothing to simulate loss of local sharpness. \\

Rician-like noise & $\sigma \in [0.02,0.20]$ & Computes $\sqrt{(x+n_1)^2+n_2^2}$ using two independent Gaussian noise fields. \\

Directional motion blur & Gaussian $\sigma \in [0.5,3.0]$ & Applies one-dimensional Gaussian smoothing along a randomly selected spatial axis. \\

Salt-and-pepper noise & Voxel probability $\in [0.02,0.20]$ & Randomly sets affected voxels to 0 or 1 with equal probability. \\

Speckle noise & $\sigma \in [0.02,0.20]$ & Applies multiplicative noise as $x(1+n)$, where $n$ is Gaussian noise. \\

Bias field & Strength $\in [0.10,0.60]$ & Generates a low-resolution random field, smooths it with Gaussian filtering, upsamples it using trilinear interpolation, normalizes it, and applies it multiplicatively as $\exp(b)$. \\

Ghosting-like artifact & Intensity $\in [0.10,0.40]$, number of ghosts $=2$--$4$ & Adds randomly shifted copies of the image along a randomly selected spatial axis. \\

Intensity shift & Shift magnitude $\in [0.05,0.40]$ & Adds or subtracts a global intensity offset with equal probability. \\

Contrast reduction & Blend factor $\in [0.10,0.60]$ & Blends the image toward its global mean intensity. \\

Resolution degradation & Downsampling factor $\in [1.5,3.0]$ & Downsamples and upsamples the image using trilinear interpolation. \\

Stripe artifact & Strength $\in [0.10,0.60]$, stripe fraction $=0.15$ & Selects random slices along a random spatial axis and adds slice-wise intensity offsets. \\

Compound corruption & Level $\in [0.03,0.15]$ & Sequentially applies Gaussian noise, Gaussian blur with $\sigma=5 \times$ level, and low-frequency multiplicative bias with strength $=3 \times$ level. \\
\hline
\end{tabular}
}
\end{table*}

\endgroup

This design reflects the assumption that real-world CT degradation is often compound rather than isolated. A clinical scan may simultaneously exhibit noise, resolution loss, altered contrast, local intensity non-uniformity, and artifact-like structures. Training under stochastic combinations of these perturbations encourages the segmentation network to learn representations that are less dependent on clean-image appearance alone and more stable under heterogeneous imaging conditions.

\subsection{Compared Augmentation Strategies}

To isolate the contribution of RAMP, we compared the proposed method against several augmentation strategies. The baseline method used the standard nnU-Net training pipeline. The intensity-focused strategy applied CT appearance perturbations such as gamma correction, window-level shift, histogram warping, local contrast adjustment, and sharpening. The geometric strategy emphasized spatial transformations, including flipping and deformation. The artifact-focused strategy applied selected artifact-like perturbations. The proposed RAMP strategy combined spatial perturbation, intensity transformation, and stochastic multi-corruption composition.

All methods used the same segmentation backbone, training schedule, preprocessing, and evaluation protocol. Therefore, differences in robustness performance can be primarily attributed to the augmentation strategy.

\subsection{Robustness Evaluation Metrics}

Segmentation performance was evaluated using the Dice similarity coefficient. In addition to clean-image Dice, we evaluated robustness under corrupted imaging conditions. Let $\mathrm{Dice}_{clean}$ denote the mean Dice score on clean images and $\mathrm{Dice}_{corr}$ denote the mean Dice score across corrupted images. We defined the clean-to-corrupted robustness gap as:
\begin{equation}
    \mathrm{Robustness\ Gap}
    =
    \mathrm{Dice}_{clean}
    -
    \mathrm{Dice}_{corr}.
\end{equation}

A smaller robustness gap indicates better stability under image degradation. We also computed the worst-case corrupted Dice:
\begin{equation}
    \mathrm{Worst\text{-}case\ Dice}
    =
    \min_{c \in \mathcal{C}_{test}} \mathrm{Dice}_{c},
\end{equation}
where $\mathcal{C}_{test}$ denotes the set of corruption types used during robustness evaluation. This metric captures whether a model collapses under severe degradation. These robustness-oriented metrics were used because clean Dice alone may overestimate clinical reliability under heterogeneous deployment conditions.

\section{Dataset}

\subsection{TotalSegmentator Subset for Five-organ CT Segmentation}

The first evaluation setting was constructed from a subset of the TotalSegmentator dataset, a large-scale CT segmentation dataset that provides voxel-level annotations for multiple anatomical structures across whole-body CT scans \cite{wasserthal2023totalsegmentator}. Rather than using all available anatomical labels, we selected a five-organ subset to create a focused robustness evaluation benchmark for clinically relevant CT segmentation. The selected target structures were the liver, spleen, kidney, heart, and colon. These organs were chosen because they represent anatomically diverse structures with different sizes, shapes, intensity distributions, and segmentation difficulties, allowing evaluation of robustness across both large high-contrast organs and more challenging structures.

For this study, the original TotalSegmentator label maps were converted into a task-specific five-organ segmentation format. Only the selected anatomical classes were retained, and all non-target structures were assigned to the background class. When the original TotalSegmentator annotations contained laterality-specific labels, such as left and right kidneys, the labels were harmonized into a single organ category when required by the experimental setting. This conversion allowed the model to be trained and evaluated as a multi-class five-organ segmentation task while preserving the voxel-level anatomical supervision provided by the original dataset.

The purpose of using this TotalSegmentator subset was to evaluate whether the proposed augmentation strategy improves segmentation robustness under controlled image degradation. TotalSegmentator is suitable for this purpose because it contains CT scans with diverse anatomical coverage and clinical imaging characteristics, making it a strong public dataset for evaluating general-purpose CT segmentation methods. However, because the present study focused on robustness rather than full-body segmentation, the task was restricted to the five selected organs instead of the complete TotalSegmentator label set.

All CT volumes were preprocessed using a consistent segmentation pipeline. CT intensities were clipped to a task-appropriate Hounsfield unit range and normalized to the interval $[0,1]$ before augmentation. Volumes were resampled, cropped, and patched according to the nnU-Net configuration used in the experiments. The same preprocessing procedure was applied to all compared augmentation strategies to ensure that performance differences were attributable to the augmentation method rather than differences in input preparation.

For robustness evaluation, corrupted test images were generated from the clean held-out TotalSegmentator subset by applying controlled image degradation operators. The corruption categories included Gaussian noise, blur, Rician-like noise, motion blur, salt-and-pepper noise, speckle noise, low-frequency intensity non-uniformity, ghosting-like artifact, global intensity shift, contrast reduction, resolution degradation, stripe or banding artifact, and compound corruption. These perturbations were intended to approximate image quality variations that may arise from scanner differences, acquisition protocols, reconstruction settings, patient motion, dose variation, and image transfer or preprocessing pipelines.

Importantly, the corruption operators were applied only to the CT images, while the corresponding segmentation labels were kept unchanged. This design enabled direct comparison between clean and degraded inputs using the same anatomical ground truth. Therefore, any reduction in Dice score under corrupted conditions reflects the sensitivity of the segmentation model to image degradation rather than differences in annotation. The TotalSegmentator subset was thus used as a controlled public benchmark for evaluating both clean segmentation accuracy and robustness under simulated clinical image degradation.

\subsection{AbdomenCT-1K Dataset}

The second evaluation setting used AbdomenCT-1K, a large-scale abdominal CT segmentation dataset introduced to evaluate whether abdominal organ segmentation can be considered a solved problem \cite{ma2022abdomenct1k}. AbdomenCT-1K contains more than 1,000 CT scans collected from multiple medical centers and includes diverse imaging conditions, vendors, phases, and disease patterns. This diversity makes it particularly useful for evaluating segmentation robustness and generalization beyond a single curated dataset.

In this study, AbdomenCT-1K was used as an additional abdominal organ segmentation benchmark. We evaluated segmentation performance on clean images and under controlled corruption conditions. The primary target organs in this evaluation included the liver and spleen. These organs were selected because they are clinically relevant for abdominal volumetry, surgical planning, and quantitative imaging analysis.

The AbdomenCT-1K images were preprocessed using the same intensity normalization and spatial preprocessing protocol used for the other CT segmentation experiments. During training, the corresponding augmentation strategy was applied only to the training images. During testing, the trained models were evaluated on both clean images and corrupted images generated from the held-out test set.

\subsection{Relationship to Existing CT Segmentation Benchmarks}

The dataset design and evaluation protocol were motivated by recent large-scale CT segmentation resources. The Medical Segmentation Decathlon demonstrated the importance of evaluating generalizable segmentation methods across multiple biomedical segmentation tasks \cite{antonelli2022msd}. TotalSegmentator further showed that robust multi-structure segmentation across whole-body CT images is feasible when models are trained on diverse clinical data \cite{wasserthal2023totalsegmentator}. However, these benchmarks primarily emphasize segmentation performance on naturally acquired or clean evaluation images. In contrast, the present study explicitly evaluates model behavior under controlled image degradation, thereby focusing on robustness as a pre-deployment requirement for clinical segmentation systems.

\subsection{Train-test Protocol}

For each dataset, the same train-test split was used across all compared augmentation methods. No test images were used during model training or augmentation parameter tuning. The corrupted test sets were generated only after training was completed and were used solely for robustness evaluation. This protocol was used to prevent information leakage and to ensure that robustness comparisons reflected the effect of training-time augmentation rather than direct optimization on the test images.

All experiments compared the following augmentation settings under identical model and training configurations: baseline nnU-Net, intensity-focused augmentation, geometric augmentation, artifact-focused augmentation, and the proposed RAMP framework. The primary endpoints were clean Dice, mean corrupted Dice, robustness gap, and worst-case corrupted Dice.

\subsection{Datasets, Splits, and Label Mapping}
\label{sec:datasets}
\begin{table*}[!htbp]
\centering
\scriptsize
\caption{Dataset composition and patient-level splits. Splits are
deterministic with seed~$42$. AbdomenCT-1K is used only as a
held-out cross-dataset evaluation set; the train/val partitions are
unused in this paper. Each subject contributes exactly one CT volume,
so patient-level and volume-level splits coincide. The mean Dice on
AbdomenCT-1K is averaged over the three shared classes
$\{$liver, spleen, kidney$\}$; pancreas, heart, and colon are reported
per-class but excluded from the AbdomenCT-1K average for the reasons
stated in Section~\ref{sec:datasets}.}
\label{tab:datasets}
\setlength{\tabcolsep}{4pt}
\renewcommand{\arraystretch}{1.2}
\begin{tabular}{l c c c c c}
\toprule
Dataset & Total subjects & Train & Validation & Test & Vols / subject \\
\midrule
TotalSegmentator (subset) & $1228$ & $859$ & $184$ & $185$ & $1$ \\
AbdomenCT-1K (cross-dataset, test only) & $1000_{\text{abd}}$ & -- & -- & $1000_{\text{abd}}^{\text{test}}$ & $1$ \\
\bottomrule
\end{tabular}
\end{table*}

\begin{table*}[!htbp]
\centering
\scriptsize
\caption{Label mapping from each dataset's native annotations to the
shared five-foreground-class output head used throughout this paper.
``--'' indicates that a class is not present in the dataset
(AbdomenCT-1K provides no heart or colon annotations; AbdomenCT-1K's
pancreas annotations are not reproducible because the trained model
has no pancreas output head). Excluded classes are still reported
per-class but do not contribute to the corresponding mean Dice.}
\label{tab:label_mapping}
\setlength{\tabcolsep}{4pt}
\renewcommand{\arraystretch}{1.2}
\begin{tabular}{l l l c}
\toprule
Output class & TotalSegmentator source label(s) & AbdomenCT-1K source ID & In AbdomenCT-1K mean \\
\midrule
liver  & \texttt{liver}                                              & 1            & Yes \\
spleen & \texttt{spleen}                                             & 3            & Yes \\
kidney & \texttt{kidney\_left} $\cup$ \texttt{kidney\_right}         & 2            & Yes \\
heart  & \texttt{heart}                                              & --           & No (absent) \\
colon  & \texttt{colon} (whole large bowel)                          & --           & No (absent) \\
\midrule
pancreas (not a model output class) & --                                  & 4            & No (no head) \\
\bottomrule
\end{tabular}
\end{table*}

\paragraph{TotalSegmentator subset.}
We use a five-organ subset of TotalSegmentator~v2 (liver, spleen, kidney,
heart, colon). After filtering for subjects whose CT volume and every
organ-of-interest mask file are present, the resulting cohort contains
\textbf{$1228$ subjects}. We partition this cohort into
\textbf{$70\%/15\%/15\%$} training/validation/test splits ($859$/$184$/$185$
subjects) using a fixed random seed ($42$). Because TotalSegmentator
provides exactly one CT volume per subject, the patient-level and
volume-level splits coincide; a single split is shared across every
ablation, leave-one-corruption-family-out, and sensitivity experiment in
this paper, so all numerical comparisons are made on the same held-out
$185$ subjects.

\paragraph{AbdomenCT-1K cross-dataset evaluation.}
AbdomenCT-1K~\cite{ma2021abdomenct} is used as an out-of-domain
benchmark; no model parameter is updated on AbdomenCT-1K data. After
matching each \texttt{images/Case\_*\_0000.nii.gz} with its mask file in
\texttt{masks/}, the cohort contains \textbf{$1000_{\text{abd}}$ subjects},
from which we hold out a test partition of $1000_{\text{abd}}^{\text{test}}$
subjects using the same $70/15/15$ convention (seed $42$); only this
test partition is used. As with TotalSegmentator, AbdomenCT-1K provides
one CT volume per subject, so the partition is patient-level.

\paragraph{Five-organ output head and label mapping.}
The model is trained with a single shared five-foreground-class output
head: liver, spleen, kidney, heart, colon (Table~\ref{tab:label_mapping}).
For TotalSegmentator, \emph{kidney} is the union of
\texttt{kidney\_left} and \texttt{kidney\_right}, while
\emph{liver}, \emph{spleen}, \emph{heart}, and \emph{colon} are
direct one-to-one mappings of the corresponding TotalSegmentator
NIfTI files. In particular, \emph{colon} is the single anatomical
TotalSegmentator label \texttt{colon} (i.e.\ the entire large bowel,
without subdivision into ascending, transverse, descending, or sigmoid
segments). For AbdomenCT-1K we follow the standard label convention
(integer ID $1$ = liver, $2$ = kidney, $3$ = spleen, $4$ = pancreas)
and remap each ID to its corresponding output class. AbdomenCT-1K masks
contain no \emph{heart} or \emph{colon} voxels, so those two classes
are absent from the AbdomenCT-1K ground truth; AbdomenCT-1K's pancreas
voxels (ID~$4$) cannot be predicted because the trained model has no
pancreas output head. Consequently, \emph{heart}, \emph{colon}, and
\emph{pancreas} are excluded from the AbdomenCT-1K mean Dice average:
the AbdomenCT-1K column reports the mean over the three shared classes
\{liver, spleen, kidney\}. We deliberately retain kidney rather than
limiting the comparison to \{liver, spleen\}, because both datasets
contain a single contiguous kidney label on each side and therefore
support a direct evaluation of kidney generalization across imaging
distributions. Per-class Dice for the three excluded classes is still
computed and reported in the experiment logs for transparency, but
does not contribute to the AbdomenCT-1K averages reported in
Table~\ref{tab:ramp_component_ablation} and elsewhere.

\paragraph{Validation set and best-checkpoint selection.}
A held-out validation partition of $184$ TotalSegmentator subjects is
used during every training run. At the end of each epoch we compute the
mean Dice over the included foreground classes on this validation set,
and retain the checkpoint with the highest validation mean Dice as the
best model. All clean-test and corruption-test results in this paper
are produced by reloading this best-validation checkpoint and
evaluating it once on the held-out test partition --- no test example
participates in either training or model selection, and no
hyperparameter (including pRAMP) is selected against the test set.

\paragraph{Corruption test protocol.}
For each held-out clean test volume we generate
$13~\text{corruption types} \times 5~\text{severity levels} = 65$
corrupted versions on-the-fly during evaluation, holding the underlying
CT volume fixed (the same $185$ TotalSegmentator test volumes
underlie all $65$ corrupted variants). The thirteen corruption types
match the families in our training-time corruption registry --- gaussian
noise, isotropic blur, Rician noise, motion blur, salt-and-pepper,
speckle, bias field, ghosting, intensity shift, contrast reduction,
resolution degradation (downsample$+$upsample), stripe/banding, and a
three-way compound (gaussian$+$blur$+$bias) --- and severity levels are
$\{0.05, 0.10, 0.15, 0.20, 0.25\}$. Mean corrupted Dice is therefore
the average over $185 \times 13 \times 5 = 12{,}025$ evaluated volumes
per training configuration; worst-case Dice is computed by taking, for
each test subject, the minimum Dice across the $65$ corruption settings
and then averaging across the test set. Clean-test Dice is reported on
the same $185$ uncorrupted volumes.

\section{Results}

\subsection{Overall Segmentation Performance Under Clean and Corrupted Conditions}

Table~\ref{tab:overall_results} summarizes the clean-image Dice score, mean corrupted Dice score, clean-to-corrupted robustness gap, and worst-case corrupted Dice score for all compared augmentation strategies. Across both evaluation settings, the proposed RAMP framework achieved the highest mean corrupted Dice score and the smallest robustness gap.

In the five-organ noisy evaluation benchmark, the standard nnU-Net baseline achieved a clean Dice score of 0.874, but its mean corrupted Dice decreased to 0.610, corresponding to a robustness gap of 0.264. In contrast, RAMP achieved a mean corrupted Dice score of 0.753 and reduced the robustness gap to 0.064. Although RAMP showed a lower clean Dice score than several conventional augmentation strategies, it substantially improved robustness under image degradation.

A similar trend was observed in AbdomenCT-1K. The nnU-Net baseline achieved a clean Dice score of 0.923 and a mean corrupted Dice score of 0.633, whereas RAMP achieved a mean corrupted Dice score of 0.789. The robustness gap was reduced from 0.290 in the nnU-Net baseline to 0.070 with RAMP. Importantly, conventional augmentation methods exhibited near-complete segmentation collapse under some severe corruption settings, with worst-case corrupted Dice scores approaching zero. RAMP preserved substantially higher worst-case Dice scores in both evaluation settings.

\begin{table*}[t]
\centering
\small
\caption{Overall segmentation performance under clean and corrupted CT imaging conditions. Mean corrupted Dice was computed across all corruption types and severity levels, excluding clean images. The robustness gap was defined as clean Dice minus mean corrupted Dice.}
\label{tab:overall_results}
\resizebox{\textwidth}{!}{
\begin{tabular}{llcccc}
\hline
Dataset & Method & Clean Dice & Mean Corrupted Dice & Robustness Gap & Worst-case Dice \\
\hline
Five-organ & nnU-Net baseline & 0.874 & 0.610 & 0.264 & 0.001 \\
Five-organ & Intensity augmentation & 0.864 & 0.662 & 0.202 & 0.086 \\
Five-organ & Geometric augmentation & \textbf{0.902} & 0.481 & 0.422 & 0.000 \\
Five-organ & Artifact augmentation & 0.889 & 0.553 & 0.336 & 0.000 \\
Five-organ & RAMP (ours) & 0.816 & \textbf{0.753} & \textbf{0.064} & \textbf{0.439} \\
\hline
AbdomenCT-1K & nnU-Net baseline & 0.923 & 0.633 & 0.290 & 0.000 \\
AbdomenCT-1K & Intensity augmentation & 0.865 & 0.620 & 0.244 & 0.016 \\
AbdomenCT-1K & Geometric augmentation & 0.925 & 0.453 & 0.471 & 0.000 \\
AbdomenCT-1K & Artifact augmentation & \textbf{0.929} & 0.521 & 0.408 & 0.000 \\
AbdomenCT-1K & RAMP (ours) & 0.859 & \textbf{0.789} & \textbf{0.070} & \textbf{0.382} \\
\hline
\end{tabular}
}
\end{table*}
\begin{figure*}[t]
    \centering
    \includegraphics[width=0.85\textwidth]{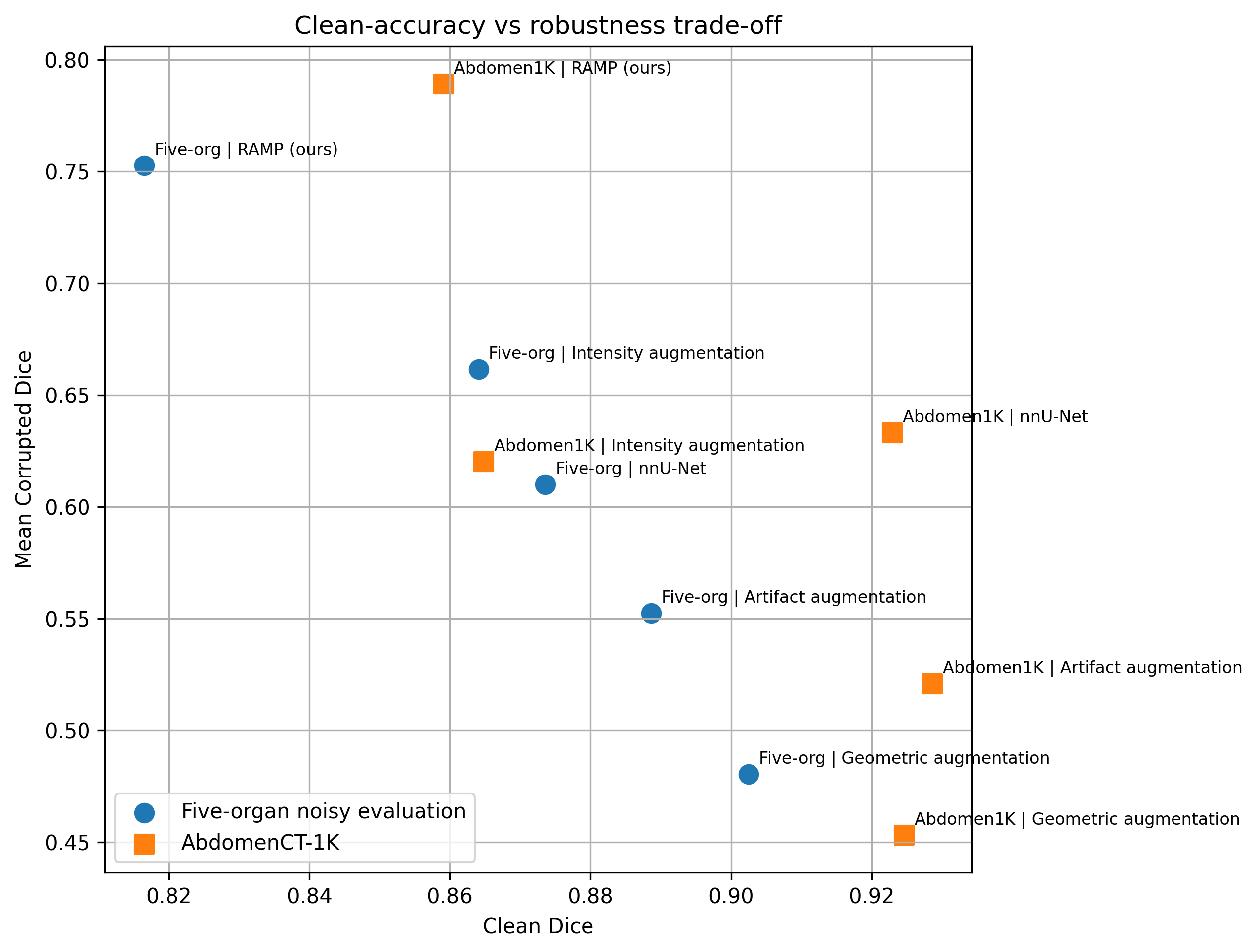}
    \caption{Clean-image Dice versus mean corrupted Dice across augmentation strategies. RAMP occupies a robustness-oriented operating point, achieving the highest corrupted-image performance despite not maximizing clean-image Dice.}
    \label{fig:clean_corrupted_tradeoff}
\end{figure*}

\subsection{RAMP Substantially Reduced the Clean-to-Corrupted Robustness Gap}

The primary objective of RAMP was to reduce performance degradation under image corruption rather than to maximize clean-image performance alone. Compared with the nnU-Net baseline, RAMP improved mean corrupted Dice by 0.143 in the five-organ noisy evaluation benchmark and by 0.156 in AbdomenCT-1K. The robustness gap was reduced by 0.200 and 0.220, respectively, corresponding to approximately 75.8\% and 75.9\% relative reductions.

Compared with the best non-RAMP augmentation strategy in the five-organ noisy evaluation benchmark, RAMP improved mean corrupted Dice by 0.091 and reduced the robustness gap by 0.138. In AbdomenCT-1K, the best non-RAMP method in terms of mean corrupted Dice was the nnU-Net baseline, and RAMP improved corrupted Dice by 0.156. These results indicate that the proposed multi-corruption training strategy provides a more stable segmentation model under heterogeneous degradation conditions.

\begin{table*}[t]
\centering
\small
\caption{Robustness improvement of RAMP compared with the nnU-Net baseline and the best non-RAMP augmentation strategy.}
\label{tab:ramp_improvement}
\begin{tabular}{llccc}
\hline
Dataset & Comparison & Mean Dice Gain & Gap Reduction & Relative Gap Reduction \\
\hline
Five-organ & RAMP vs. nnU-Net & +0.143 & 0.200 & 75.8\% \\
Five-organ & RAMP vs. best non-RAMP & +0.091 & 0.138 & -- \\
AbdomenCT-1K & RAMP vs. nnU-Net & +0.156 & 0.220 & 75.9\% \\
AbdomenCT-1K & RAMP vs. best non-RAMP & +0.156 & 0.220 & -- \\
\hline
\end{tabular}
\end{table*}

\subsection{Corruption-wise Robustness Analysis}

To further characterize robustness behavior, we analyzed performance separately for each corruption family. Tables~\ref{tab:corruptionwise_fiveorgan} and~\ref{tab:corruptionwise_abdomenct1k} compare RAMP with the best non-RAMP method for each corruption type. RAMP showed the largest advantages under corruptions that caused severe model instability in conventional methods, including bias field, compound corruption, Gaussian noise, Rician-like noise, salt-and-pepper noise, and stripe artifacts.

In the five-organ noisy evaluation benchmark, RAMP outperformed the best non-RAMP method for bias field, compound corruption, Gaussian noise, Rician-like noise, salt-and-pepper noise, and stripe artifacts. The largest gain was observed for salt-and-pepper noise, where RAMP improved the corruption-wise mean Dice by 0.409 over the best conventional augmentation strategy. RAMP also improved compound corruption robustness by 0.290 and stripe artifact robustness by 0.188.

In AbdomenCT-1K, RAMP again demonstrated strong gains under severe and destabilizing corruption families. RAMP improved bias field robustness by 0.326, compound corruption robustness by 0.286, Gaussian noise robustness by 0.155, Rician-like noise robustness by 0.145, salt-and-pepper robustness by 0.507, and stripe artifact robustness by 0.392. These results suggest that RAMP is particularly effective at preventing segmentation collapse under corruption types that strongly disrupt image intensity, local texture, and spatial consistency.

\begin{table}[t]
\centering
\scriptsize
\caption{Corruption-wise mean Dice comparison between RAMP and the best non-RAMP method in the five-organ noisy evaluation benchmark. Each value represents the mean Dice across severity levels for the corresponding corruption type.}
\label{tab:corruptionwise_fiveorgan}
\begin{tabular}{lccc}
\hline
Corruption Type & Best non-RAMP Dice & RAMP Dice & Difference \\
\hline
Bias field & 0.621 & \textbf{0.760} & +0.139 \\
Blur & \textbf{0.863} & 0.776 & -0.087 \\
Compound & 0.379 & \textbf{0.669} & +0.290 \\
Contrast & \textbf{0.830} & 0.800 & -0.030 \\
Downsample & \textbf{0.870} & 0.807 & -0.063 \\
Gaussian noise & 0.610 & \textbf{0.700} & +0.090 \\
Ghosting & \textbf{0.838} & 0.798 & -0.040 \\
Intensity shift & \textbf{0.788} & 0.766 & -0.022 \\
Motion blur & \textbf{0.876} & 0.797 & -0.079 \\
Rician-like noise & 0.554 & \textbf{0.649} & +0.095 \\
Salt-and-pepper noise & 0.293 & \textbf{0.702} & +0.409 \\
Speckle noise & \textbf{0.877} & 0.791 & -0.086 \\
Stripe artifact & 0.582 & \textbf{0.770} & +0.188 \\
\hline
\end{tabular}
\end{table}

\begin{table}[t]
\centering
\scriptsize
\caption{Corruption-wise mean Dice comparison between RAMP and the best non-RAMP method in AbdomenCT-1K. Each value represents the mean Dice across severity levels for the corresponding corruption type.}
\label{tab:corruptionwise_abdomenct1k}
\begin{tabular}{lccc}
\hline
Corruption Type & Best non-RAMP Dice & RAMP Dice & Difference \\
\hline
Bias field & 0.495 & \textbf{0.821} & +0.326 \\
Blur & \textbf{0.911} & 0.847 & -0.064 \\
Compound & 0.356 & \textbf{0.642} & +0.286 \\
Contrast & \textbf{0.857} & 0.847 & -0.010 \\
Downsample & \textbf{0.925} & 0.857 & -0.068 \\
Gaussian noise & 0.570 & \textbf{0.725} & +0.155 \\
Ghosting & \textbf{0.837} & 0.811 & -0.026 \\
Intensity shift & 0.748 & \textbf{0.778} & +0.030 \\
Motion blur & \textbf{0.917} & 0.849 & -0.068 \\
Rician-like noise & 0.506 & \textbf{0.651} & +0.145 \\
Salt-and-pepper noise & 0.253 & \textbf{0.760} & +0.507 \\
Speckle noise & \textbf{0.918} & 0.840 & -0.078 \\
Stripe artifact & 0.441 & \textbf{0.833} & +0.392 \\
\hline
\end{tabular}
\end{table}

\begin{figure*}[t]
    \centering
    \includegraphics[width=\textwidth]{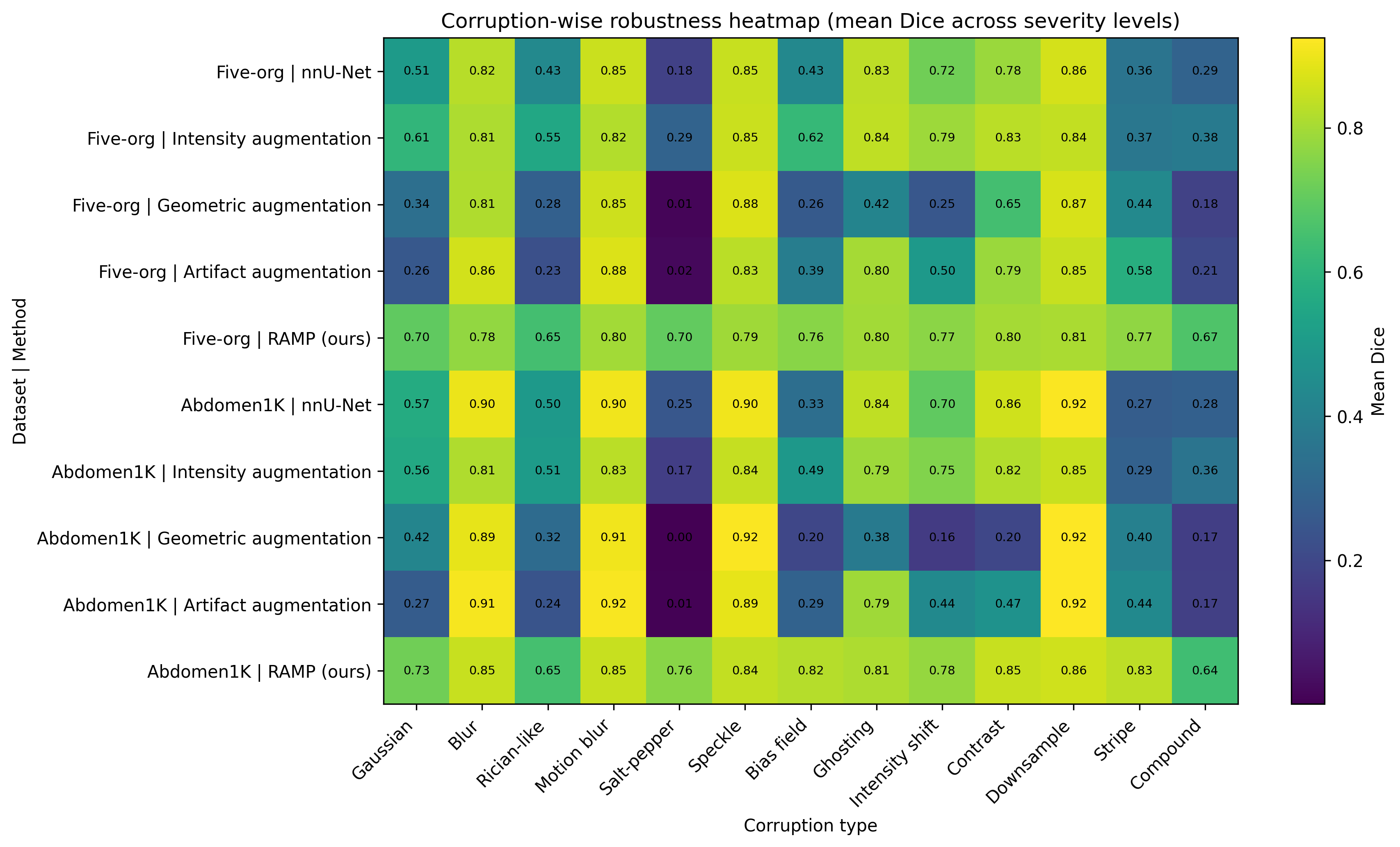}
    \caption{Corruption-wise robustness heatmap across augmentation strategies. Each cell represents mean Dice across severity levels for the corresponding corruption type. RAMP shows the strongest robustness under severe intensity- and artifact-related corruptions, including bias field, compound corruption, salt-and-pepper noise, and stripe artifacts.}
    \label{fig:corruption_heatmap}
\end{figure*}

\subsection{Robustness Under High-Severity Corruptions}

We next examined performance under high-severity degradation, defined as corruption levels greater than or equal to 0.15. This analysis was intended to evaluate whether each augmentation strategy could prevent segmentation collapse under challenging image quality conditions.

In the five-organ noisy evaluation benchmark, RAMP achieved the highest Dice score in 21 of 39 high-severity corruption settings. In AbdomenCT-1K, RAMP achieved the highest Dice score in 23 of 39 high-severity corruption settings. These results indicate that the advantage of RAMP becomes particularly apparent under severe degradation, where conventional augmentation strategies often fail to preserve stable segmentation performance.

\begin{table*}[t]
\centering
\small
\caption{Number of high-severity corruption settings in which each method achieved the highest Dice score. High-severity corruption was defined as severity level greater than or equal to 0.15.}
\label{tab:high_severity_winners}
\begin{tabular}{lccccc}
\hline
Dataset & nnU-Net & Intensity & Geometric & Artifact & RAMP \\
\hline
Five-organ & 1 & 6 & 5 & 6 & \textbf{21} \\
AbdomenCT-1K & 4 & 0 & 3 & 9 & \textbf{23} \\
\hline
\end{tabular}
\end{table*}

\begin{figure*}[t]
    \centering
    \includegraphics[width=\textwidth]{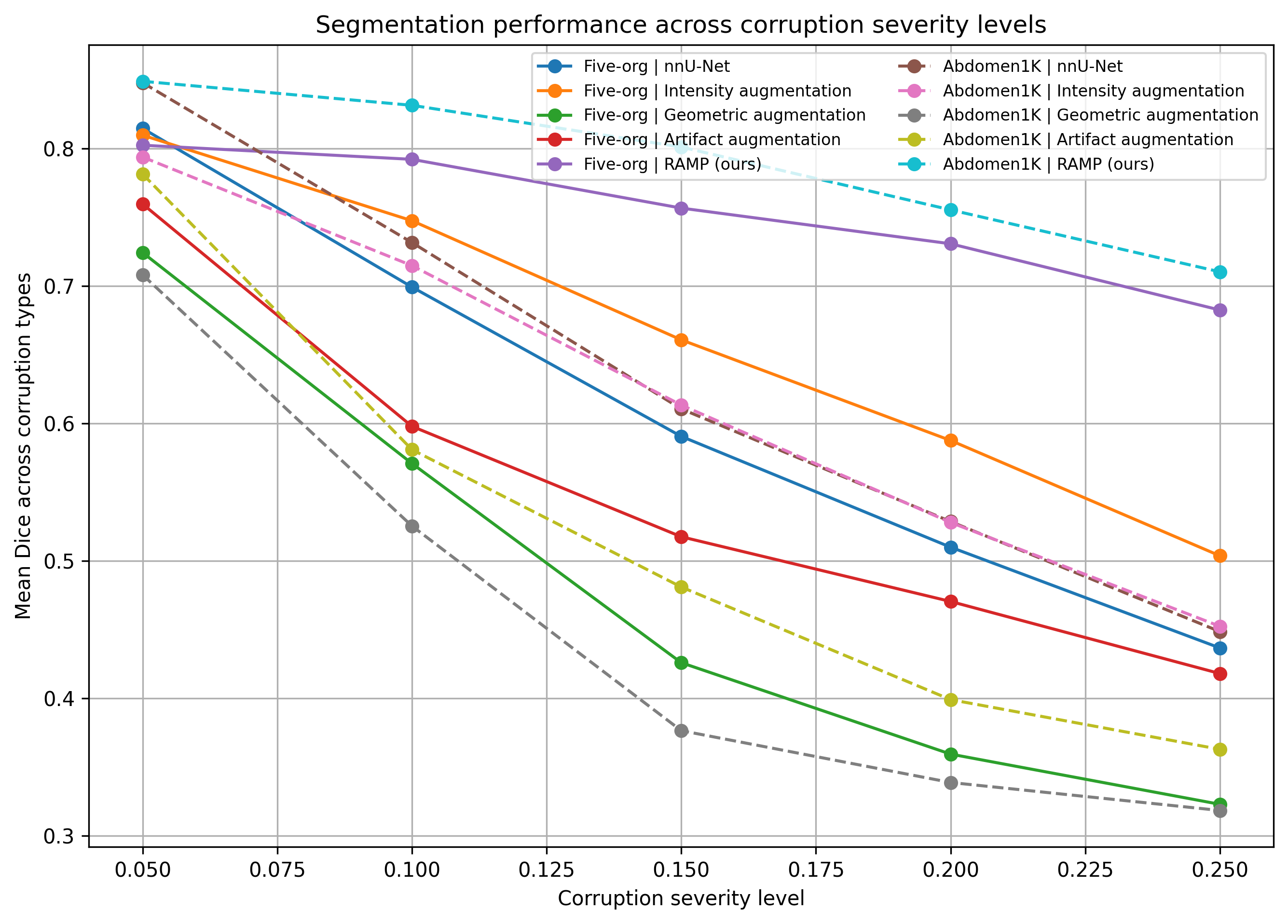}
    \caption{Segmentation performance across corruption severity levels. RAMP shows a slower degradation pattern than conventional augmentation strategies, indicating improved stability under increasing image degradation.}
    \label{fig:severity_curve}
\end{figure*}

\subsection{Organ-wise Robustness of RAMP}

Table~\ref{tab:organwise_ramp} summarizes organ-wise clean Dice, mean corrupted Dice, robustness gap, and worst-case corrupted Dice for RAMP. In the five-organ noisy evaluation benchmark, RAMP maintained relatively small robustness gaps for the liver, kidney, heart, and colon. The spleen showed a larger robustness gap and lower worst-case Dice, suggesting that smaller or more appearance-sensitive organs may remain more vulnerable under severe degradation. The heart showed the most stable performance, with a clean Dice of 0.897, mean corrupted Dice of 0.861, and robustness gap of 0.036.

In AbdomenCT-1K, RAMP showed similar organ-wise behavior. The liver maintained a clean Dice of 0.910 and a mean corrupted Dice of 0.858, corresponding to a robustness gap of 0.052. The spleen had a lower mean corrupted Dice of 0.720 and a larger robustness gap of 0.087. These findings suggest that RAMP improves overall robustness, but residual organ-specific vulnerability remains, particularly for structures that are more sensitive to image degradation.

\begin{table*}[]
\centering
\small
\caption{Organ-wise robustness of RAMP. Mean corrupted Dice was computed across all corruption types and severity levels, excluding clean images.}
\label{tab:organwise_ramp}
\begin{tabular}{llcccc}
\hline
Dataset & Organ & Clean Dice & Corrupted Dice & Gap & Worst-case Dice \\
\hline
Five-organ & Liver & 0.908 & 0.857 & 0.052 & 0.590 \\
Five-organ & Spleen & 0.823 & 0.711 & 0.112 & 0.217 \\
Five-organ & Kidney & 0.837 & 0.761 & 0.076 & 0.350 \\
Five-organ & Heart & 0.897 & 0.861 & 0.036 & 0.745 \\
Five-organ & Colon & 0.617 & 0.573 & 0.044 & 0.244 \\
\hline
AbdomenCT-1K & Liver & 0.910 & 0.858 & 0.052 & 0.582 \\
AbdomenCT-1K & Spleen & 0.808 & 0.720 & 0.087 & 0.181 \\
\hline
\end{tabular}
\end{table*}

\subsection{Clean Accuracy Versus Robustness Trade-off}

Although RAMP achieved the strongest robustness profile, it did not achieve the highest clean-image Dice score. In the five-organ noisy evaluation benchmark, geometric augmentation achieved the highest clean Dice score of 0.902, whereas RAMP achieved a clean Dice score of 0.816. In AbdomenCT-1K, artifact augmentation achieved the highest clean Dice score of 0.929, whereas RAMP achieved 0.859.

This finding indicates a clear clean-robustness trade-off. Conventional augmentation strategies can preserve or improve clean-image performance but may remain vulnerable to severe image degradation. In contrast, RAMP sacrifices some clean-image accuracy in exchange for substantially improved corrupted-image performance, smaller robustness gaps, and higher worst-case Dice scores. For deployment-oriented medical imaging systems, this trade-off may be clinically meaningful because stable performance under heterogeneous imaging conditions is critical for reliable downstream analysis.

Overall, these results support the use of RAMP as a robustness-oriented pre-deployment strategy for CT segmentation systems. Rather than serving as a clean-image performance optimizer, RAMP provides a more stable operating point for segmentation models expected to encounter heterogeneous image quality, scanner variability, and acquisition-related degradation in real-world clinical environments.

\subsection{Additional Ablation, Generalization, and Sensitivity Analyses}

To further characterize the behavior of RAMP, we conducted additional experiments beyond the primary robustness comparison. These analyses were designed to answer four methodological questions. First, we examined which components of RAMP were responsible for the observed clean-performance trade-off. Second, we evaluated whether RAMP generalized to corruption families that were intentionally excluded during training. Third, we assessed whether increasing the number of composed corruptions changed segmentation performance. Fourth, we examined the sensitivity of RAMP to the probability of applying multi-corruption augmentation.

All additional analyses were performed on both the TotalSegmentator subset and AbdomenCT-1K. The TotalSegmentator subset included five target organs: liver, spleen, kidney, heart, and colon. In AbdomenCT-1K, the reported Dice scores were computed over the included abdominal organs, while excluded classes were not included in the mean Dice calculation. The purpose of these experiments was not to identify the configuration with the highest clean-image Dice score, but to understand how RAMP controls the trade-off between clean-image accuracy and robustness-oriented augmentation strength.

\subsubsection{Component Ablation of RAMP}

RAMP consists of three major components: spatial perturbation, intensity transformation, and stochastic multi-corruption composition. To evaluate the contribution of each component, we trained models using seven augmentation settings: baseline training, spatial-only augmentation, intensity-only augmentation, corruption-only augmentation, spatial plus intensity augmentation, intensity plus corruption augmentation, and full RAMP. The results are summarized in Table~\ref{tab:ramp_component_ablation}.

On the TotalSegmentator subset, the baseline model achieved the highest clean-image mean Dice score of 0.938. Intensity-only augmentation achieved a similar Dice score of 0.930, suggesting that CT appearance transformations such as gamma correction, window-level shift, histogram warping, local contrast adjustment, and sharpening did not substantially degrade clean-image segmentation performance. In contrast, corruption-only augmentation reduced the mean Dice to 0.886, and intensity plus corruption augmentation achieved 0.878. Full RAMP, which combines spatial, intensity, and multi-corruption augmentation, achieved a mean Dice of 0.786.

A similar pattern was observed in AbdomenCT-1K. The baseline model achieved the highest mean Dice score of 0.954, followed closely by intensity-only augmentation with 0.952. Corruption-only augmentation achieved 0.938, intensity plus corruption achieved 0.930, and full RAMP achieved 0.868. These findings indicate that the reduction in clean-image Dice is primarily associated with the corruption exposure component, rather than intensity augmentation alone.

The ablation results therefore clarify the role of RAMP. Full RAMP is not designed to maximize clean-image Dice. Instead, it applies a more aggressive robustness-oriented training regime that exposes the model to degraded and compound imaging conditions. This explains why full RAMP may show lower clean-image Dice while achieving stronger corrupted-image performance in the primary robustness evaluation.

\begin{table*}[!htbp]
\centering
\scriptsize
\caption{Component ablation of RAMP. Values represent mean Dice scores for each training configuration. Spatial, intensity, and multi-corruption columns indicate whether each component was enabled during training.}
\label{tab:ramp_component_ablation}
\setlength{\tabcolsep}{3.5pt}
\renewcommand{\arraystretch}{1.15}
\begin{tabular}{lccc cc}
\toprule
Method & Spatial & Intensity & Multi-corr. & TotalSeg & AbdomenCT-1K \\
\midrule
Baseline & -- & -- & -- & \textbf{0.938} & \textbf{0.954} \\
Spatial only & Yes & -- & -- & 0.908 & 0.937 \\
Intensity only & -- & Yes & -- & 0.930 & 0.952 \\
Corruption only & -- & -- & Yes & 0.886 & 0.938 \\
Spatial + intensity & Yes & Yes & -- & 0.890 & 0.920 \\
Intensity + corruption & -- & Yes & Yes & 0.878 & 0.930 \\
Full RAMP & Yes & Yes & Yes & 0.786 & 0.868 \\
\bottomrule
\end{tabular}
\end{table*}

\subsubsection{Interpretation of the Component Ablation}

The component ablation demonstrates a clear clean-performance hierarchy. Intensity-only augmentation preserved clean Dice most effectively, whereas corruption-based augmentation reduced clean Dice. This is expected because corruption augmentation exposes the model to training images that are intentionally degraded and therefore more difficult than clean images. The resulting model is less optimized for ideal clean-image appearance, but is expected to be more stable under degraded imaging conditions.

The difference between corruption-only and full RAMP also provides useful insight. In TotalSegmentator, corruption-only augmentation achieved a mean Dice of 0.886, while full RAMP achieved 0.786. In AbdomenCT-1K, corruption-only achieved 0.938, while full RAMP achieved 0.868. This suggests that the combination of spatial perturbation, intensity transformation, and multi-corruption exposure produces a stronger regularization effect than corruption exposure alone. Therefore, full RAMP should be interpreted as an aggressive robustness-oriented configuration rather than a conventional clean-performance augmentation strategy.

These results support the central framing of this study: RAMP is intended to provide a robustness-oriented operating point for CT segmentation systems. The clean Dice reduction observed in the component ablation should be interpreted together with the main corrupted-image evaluation, where RAMP reduced the robustness gap and improved worst-case segmentation performance.

\subsubsection{Leave-One-Corruption-Family-Out Generalization}

One potential concern with corruption-based augmentation is that the model may simply learn to handle the same corruption operators that appear during evaluation. To address this concern, we performed leave-one-corruption-family-out (LOCO) experiments. In each LOCO experiment, one corruption family was excluded from the RAMP training pool. The trained model was then evaluated on the corresponding held-out corruption family. This design tests whether the robustness benefit of RAMP extends beyond direct exposure to the exact corruption operator used during evaluation.

The LOCO experiments were conducted for eight corruption families: bias field, compound corruption, Gaussian noise, motion blur, resolution degradation, Rician-like noise, salt-and-pepper noise, and stripe artifact. The results are shown in Table~\ref{tab:loco_results}. On the TotalSegmentator subset, LOCO-RAMP maintained mean Dice scores between 0.781 and 0.818 across held-out corruption families. The highest Dice score was observed when compound corruption was excluded from training, with a mean Dice of 0.818. The lowest Dice score was observed when motion blur was excluded, with a mean Dice of 0.781.

In AbdomenCT-1K, LOCO performance was more stable across corruption families. Mean Dice scores ranged from 0.871 to 0.896. The lowest Dice score was observed for resolution degradation leave-out, whereas the highest Dice score was observed for stripe artifact leave-out. Importantly, performance did not collapse when any individual corruption family was excluded from training.

These results suggest that RAMP does not rely solely on memorization of individual corruption operators. Even when the evaluated corruption family was absent during training, the model retained meaningful segmentation performance. This supports the interpretation that RAMP improves robustness by encouraging more general tolerance to heterogeneous image degradation, rather than by simply matching training corruptions to test corruptions.

\begin{table}[!htbp]
\centering
\scriptsize
\caption{Leave-one-corruption-family-out generalization. For each row, the listed corruption family was excluded during RAMP training and used as the held-out corruption family for evaluation.}
\label{tab:loco_results}
\setlength{\tabcolsep}{4pt}
\renewcommand{\arraystretch}{1.15}
\begin{tabular}{lcc}
\toprule
Held-out corruption family & TotalSeg Dice & AbdomenCT-1K Dice \\
\midrule
Bias field & 0.782 & 0.883 \\
Compound corruption & \textbf{0.818} & 0.888 \\
Gaussian noise & 0.803 & 0.886 \\
Motion blur & 0.781 & 0.879 \\
Resolution degradation & 0.806 & 0.871 \\
Rician-like noise & 0.816 & 0.888 \\
Salt-and-pepper noise & 0.785 & 0.895 \\
Stripe artifact & 0.788 & \textbf{0.896} \\
\bottomrule
\end{tabular}
\end{table}

\FloatBarrier

\subsubsection{Effect of the Number of Composed Corruptions}

RAMP is based on the assumption that clinical image degradation is often compound rather than isolated. To examine the effect of corruption composition strength, we varied the number of corruption operators applied per image. We compared fixed single-corruption training ($k=1$), fixed multi-corruption training ($k=2$, $k=3$, and $k=4$), and the stochastic $k=2$--$4$ setting used as the main RAMP configuration. The results are summarized in Table~\ref{tab:k_ablation}.

In both datasets, $k=1$ achieved the highest Dice score among the tested settings. On the TotalSegmentator subset, $k=1$ achieved a mean Dice of 0.857. Increasing the number of corruptions reduced clean Dice: $k=2$ achieved 0.833, $k=3$ achieved 0.756, and $k=4$ achieved 0.733. The stochastic $k=2$--$4$ setting achieved 0.804, which was lower than $k=1$ and $k=2$, but higher than fixed $k=3$ and $k=4$.

In AbdomenCT-1K, $k=1$ again achieved the highest mean Dice score of 0.915. The fixed multi-corruption settings achieved lower scores: 0.878 for $k=2$, 0.877 for $k=3$, and 0.875 for $k=4$. The stochastic $k=2$--$4$ setting achieved 0.885, which was higher than the fixed $k=2$, $k=3$, and $k=4$ settings.

These findings indicate that increasing the number of composed corruptions increases the difficulty of the training distribution and can reduce clean-image Dice. However, the stochastic $k=2$--$4$ setting appears to provide a more balanced configuration than always applying a fixed high number of corruptions. By randomly varying the number of corruptions, the model is exposed to both moderate and severe degradation conditions, avoiding the excessive difficulty of consistently applying three or four corruptions to every augmented image.

\begin{table}[!htbp]
\centering
\scriptsize
\caption{Effect of the number of composed corruptions. The stochastic $k=2$--$4$ setting corresponds to the main RAMP configuration.}
\label{tab:k_ablation}
\setlength{\tabcolsep}{5pt}
\renewcommand{\arraystretch}{1.15}
\begin{tabular}{lcc}
\toprule
Number of corruptions & TotalSeg Dice & AbdomenCT-1K Dice \\
\midrule
$k=1$ & \textbf{0.857} & \textbf{0.915} \\
$k=2$ & 0.833 & 0.878 \\
$k=3$ & 0.756 & 0.877 \\
$k=4$ & 0.733 & 0.875 \\
$k=2$--$4$ & 0.804 & 0.885 \\
\bottomrule
\end{tabular}
\end{table}

\FloatBarrier

\subsubsection{Interpretation of the k-ablation}

The k-ablation results should be interpreted as a clean-robustness trade-off analysis rather than a clean-performance optimization experiment. Single-corruption training preserves clean Dice better because each augmented image remains closer to the original clean distribution. In contrast, multi-corruption training generates more difficult images that may deviate substantially from the clean training distribution. This can reduce clean-image Dice but may improve the model's tolerance to degraded imaging conditions.

The stochastic $k=2$--$4$ setting was selected as the main RAMP configuration because it reflects the assumption that real-world clinical CT degradation can vary in complexity. Some scans may contain only mild or isolated degradation, whereas others may contain multiple simultaneous sources of degradation. Randomly sampling between two and four corruptions exposes the model to a spectrum of degradation complexity. The fact that $k=2$--$4$ preserved higher Dice than fixed $k=3$ or $k=4$ in both datasets suggests that stochastic composition may be preferable to a consistently aggressive corruption setting.

\subsubsection{Sensitivity to RAMP Application Probability}

We further evaluated the sensitivity of RAMP to the probability of applying stochastic multi-corruption composition, denoted as $p_{\mathrm{RAMP}}$. This parameter controls how frequently the multi-corruption sampler is applied during training. A smaller value corresponds to weaker robustness augmentation, whereas a larger value corresponds to more aggressive exposure to degraded images. We compared three settings: $p_{\mathrm{RAMP}}=0.30$, $p_{\mathrm{RAMP}}=0.50$, and $p_{\mathrm{RAMP}}=0.85$.

The results are shown in Table~\ref{tab:pramp_sensitivity}. On the TotalSegmentator subset, increasing $p_{\mathrm{RAMP}}$ reduced clean-image Dice. The mean Dice score was 0.854 for $p_{\mathrm{RAMP}}=0.30$, 0.813 for $p_{\mathrm{RAMP}}=0.50$, and 0.804 for $p_{\mathrm{RAMP}}=0.85$. This pattern indicates that more frequent corruption exposure acts as a stronger regularizer and can reduce clean-image performance.

In AbdomenCT-1K, the Dice scores were relatively stable across the tested range. The mean Dice was 0.890 for $p_{\mathrm{RAMP}}=0.30$, 0.893 for $p_{\mathrm{RAMP}}=0.50$, and 0.897 for $p_{\mathrm{RAMP}}=0.85$. Unlike the TotalSegmentator subset, the highest clean Dice was observed at the most aggressive setting. This suggests that the effect of $p_{\mathrm{RAMP}}$ may depend on dataset characteristics, organ composition, and the degree of baseline heterogeneity in the training data.

Overall, the $p_{\mathrm{RAMP}}$ sensitivity analysis indicates that RAMP can be tuned to select different clean-robustness operating points. When clean-image performance is prioritized, lower values of $p_{\mathrm{RAMP}}$ may be preferable. When robustness under degraded imaging conditions is prioritized, higher values may be appropriate. Therefore, $p_{\mathrm{RAMP}}$ should be interpreted as a controllable robustness hyperparameter rather than a universally fixed value.

\begin{table}[!htbp]
\centering
\scriptsize
\caption{Sensitivity analysis of the RAMP application probability.}
\label{tab:pramp_sensitivity}
\setlength{\tabcolsep}{6pt}
\renewcommand{\arraystretch}{1.15}
\begin{tabular}{lcc}
\toprule
$p_{\mathrm{RAMP}}$ & TotalSeg Dice & AbdomenCT-1K Dice \\
\midrule
0.30 & \textbf{0.854} & 0.890 \\
0.50 & 0.813 & 0.893 \\
0.85 & 0.804 & \textbf{0.897} \\
\bottomrule
\end{tabular}
\end{table}

\FloatBarrier

\subsubsection{Integrated Interpretation of the Additional Experiments}

Taken together, the additional experiments support the interpretation of RAMP as a controllable robustness-oriented augmentation framework. The component ablation showed that intensity augmentation alone preserved clean-image Dice, whereas corruption-based augmentation reduced clean Dice. This indicates that the robustness-oriented component of RAMP is also the main source of clean-performance trade-off.

The LOCO experiments further addressed the concern that RAMP might simply memorize the corruption families used during evaluation. Across two datasets and eight held-out corruption families, LOCO-RAMP maintained non-trivial Dice scores even when the evaluated corruption family was excluded from training. This suggests that RAMP encourages more general robustness to image degradation rather than merely learning corruption-specific shortcuts.

The k-ablation and $p_{\mathrm{RAMP}}$ sensitivity experiments showed that the strength of corruption exposure can be controlled. Increasing the number of composed corruptions or increasing the probability of applying RAMP generally produces a stronger augmentation regime. This may reduce clean-image Dice, but it also provides a mechanism for selecting a robustness-oriented operating point depending on the intended deployment setting.

These findings are important for clinical deployment. A CT segmentation system used in a controlled research environment may prioritize clean-image Dice, whereas a model intended for heterogeneous clinical deployment may require greater tolerance to image degradation, scanner variation, reconstruction differences, and artifact patterns. RAMP provides a practical mechanism for navigating this trade-off before deployment.

Overall, the additional experiments strengthen the main conclusion of this study. RAMP should not be viewed as a conventional augmentation strategy designed only to maximize clean benchmark performance. Instead, it is a pre-deployment robustness framework that exposes segmentation models to heterogeneous and compound image degradation, supports generalization to unseen corruption families, and provides tunable parameters for controlling the clean-robustness trade-off.

\section{Limitations}

This study has several limitations. First, the corrupted evaluation sets were generated using simulated image degradation rather than prospectively collected low-quality or artifact-affected clinical CT scans. Although the corruption operators were selected to approximate plausible variations in image quality, acquisition protocol, reconstruction, resolution, contrast, and artifacts, they cannot fully represent the complexity of real-world CT imaging environments. Clinical artifacts may arise from scanner-specific reconstruction pipelines, patient motion patterns, metal implants, contrast timing, dose modulation, and institutional protocols that are difficult to reproduce using synthetic perturbations alone.

Second, RAMP did not achieve the highest clean-image Dice score. The proposed framework was designed to improve robustness under heterogeneous imaging degradation, and the results showed a clear clean-robustness trade-off. While RAMP substantially improved mean corrupted Dice, reduced robustness gaps, and prevented worst-case segmentation collapse, it achieved lower clean Dice than several conventional augmentation strategies. Therefore, RAMP should be interpreted as a robustness-oriented training strategy rather than a clean-image performance optimizer. In clinical deployment, the preferred operating point may depend on whether the application prioritizes peak performance under ideal imaging conditions or stable performance under heterogeneous image quality.

Third, the present study evaluated robustness primarily using Dice-based segmentation metrics. Although Dice score is widely used in medical image segmentation, it may not fully capture the clinical impact of segmentation errors. For example, small boundary errors may have limited clinical relevance in some volumetric tasks, whereas small errors near critical anatomical structures may be more consequential. Future work should evaluate downstream clinical endpoints, including volumetric measurement error, radiomics feature stability, surgical planning consistency, and reader- or workflow-level impact.

Fourth, the evaluation was limited to selected CT segmentation settings and anatomical structures. Although the experiments included two evaluation settings and multiple organ targets, additional validation is needed across broader anatomical regions, imaging phases, scanner vendors, institutions, and disease populations. External multicenter validation using naturally heterogeneous clinical CT data would further strengthen the evidence that RAMP improves deployment robustness beyond controlled corruption benchmarks.

Finally, this study evaluated RAMP using a fixed segmentation framework. nnU-Net was selected as a strong and reproducible baseline, but the effect of RAMP may vary across architectures, loss functions, training schedules, and preprocessing pipelines. Future work should examine whether the same robustness benefits generalize to transformer-based segmentation models, foundation models for medical imaging, and other deployment-oriented segmentation systems.

Future clinical translation studies should also follow established reporting guidance for medical imaging AI and clinical AI evaluation, including transparent documentation of data provenance, preprocessing, failure modes, deployment setting, and intended clinical use \cite{mongan2020claim,vasey2022decideai}.

\section{Conclusion}

In this study, we proposed RAMP, a robustness-oriented multi-corruption augmentation framework for CT segmentation systems. Unlike conventional augmentation strategies that primarily aim to improve clean-image performance, RAMP was designed to reduce segmentation instability under heterogeneous imaging degradation. By combining anatomically constrained spatial perturbations, CT intensity transformations, and stochastic multi-corruption composition, RAMP exposes segmentation models to a broad perturbation space during training.

Across two CT segmentation evaluation settings, RAMP achieved the strongest corrupted-image performance, the smallest clean-to-corrupted robustness gap, and the highest worst-case corrupted Dice. Although RAMP did not achieve the best clean-image Dice score, it substantially reduced performance degradation under severe corruption and prevented the near-complete segmentation collapse observed in several baseline and conventional augmentation models. These findings demonstrate that clean benchmark accuracy alone may be insufficient for assessing the reliability of CT segmentation systems intended for clinical deployment.

Overall, RAMP provides a practical pre-deployment strategy for improving the robustness of CT segmentation models under heterogeneous imaging conditions. The results support a broader evaluation paradigm in which medical AI systems are assessed not only by their performance on clean curated datasets, but also by their stability under clinically plausible image degradation. Such robustness-oriented training and stress testing may help improve the reliability, safety, and clinical usability of automated CT segmentation systems in real-world medical imaging workflows.


\bibliographystyle{cas-model2-names}

\bibliography{cas-refs}



\end{document}